\documentclass{article}
\usepackage{style}

\makeatletter
\def\hlinew#1{%
  \noalign{\ifnum0=`}\fi\hrule \@height #1 \futurelet
   \reserved@a\@xhline}
\makeatother

\usepackage{times}
\usepackage{xcolor}
\usepackage{soul}
\usepackage[utf8]{inputenc}
\usepackage[small]{caption}
\usepackage{multirow}
\usepackage{booktabs}

\usepackage{amsfonts}
\usepackage{amsmath}
\usepackage{graphicx} 
\graphicspath{{figures/}}

\newcommand{\tabincell}[2]{\begin{tabular}{@{}#1@{}}#2\end{tabular}}

\title{Graph-Adaptive Pruning for Efficient Inference of \\ Convolutional Neural Networks
}

\author{
Mengdi Wang, 
Qing Zhang, 
Jun Yang,
Xiaoyuan Cui,
Wei Lin
\\ 
Alibaba inc. \\
{\{didou.wmd, sensi.zq, muzhuo.yj, xiaoyuan.cui, weilin.lw\}@alibaba-inc.com
}
}

\begin{document}

\maketitle

\begin{abstract}
In this work, we propose a graph-adaptive pruning (GAP) method for efficient inference of convolutional neural networks (CNNs). 
In this method, the network is viewed as a computational graph, 
in which the vertices denote the computation nodes and edges represent the information flow.  Through topology analysis, GAP is capable of adapting to different network structures, especially the widely used cross connections and multi-path data flow in recent novel convolutional models. 
The models can be adaptively pruned at vertex-level as well as edge-level 
without any post-processing,  
thus GAP can directly get practical model compression and inference speed-up. Moreover, it does not need any customized computation library or hardware support.
Finetuning is conducted after pruning to restore the model performance. In the finetuning step, we adopt a self-taught knowledge distillation (KD) strategy by utilizing information from the original model, through which, the performance of the optimized model can be sufficiently improved, without introduction of any other teacher model. Experimental results show the proposed GAP can achieve promising result to make inference more efficient, e.g., for ResNeXt-29 on CIFAR10, it can get 13$\times$ model compression and 4.3$\times$ practical speed-up with marginal loss of accuracy.

\end{abstract}

\section{Introduction}


In recent years, more and more methods are proposed for CNN inference optimization, such as model quantization~\cite{gong2014compressing,han2015learning}, tensor decomposition (TD)~\cite{kim2015compression}, weight pruning~\cite{wen2016learning,liu2017learning}, knowledge distillation~\cite{hinton2015distilling,yim2017gift} and new network architecture designs~\cite{iandola2016squeezenet,howard2017mobilenets}, \emph{etc}. Most existing methods can achieve promising model compression and floating-point operations (FLOPs) reduction with marginal accuracy degradation. 
However, smaller model size and FLOPs cannot guarantee the practical speed-up for inference.
In this work, we propose the GAP method to prune the weight in order to achieve better practical inference optimization. The pruned network could be supported by any off-the-shelf deep learning libraries, thus practical acceleration can be achieved without any effort to build additional compute libraries. Our method falls into the type of weight pruning based on structural sparsity constraint.

Most existing structural sparsity-induced weight pruning methods prune the network at either channel-level or kernel-level. \cite{wen2016learning} uses group sparsity regularization to help select the removable kernels or channels, and network slimming (NS) \cite{liu2017learning}  alters to use $\ell_1$-norm on the scaling factors of batch normalization (BN) layers for channel selection. 
However, recent models are of much more complex structures, such as cross-connection, including 1-to-n connection and n-to-1 connection. For such structure, pruning a channel 
cannot result in pruning of the preceding kernel since output of the filter may still be used by other layers. As a result, post-processing may be necessary after channel-level pruning to maintain the network topology. For example, DenseNet \cite{huang2016densely} has 1-to-n connections, as the output of one layer will be reused by all the following convolution layers, and each re-usage has its own BN layer. 
After channel-level pruning, the NS method chooses to remain all the kernels before the 1-to-n connection and insert a selection layer before each connected subsequent layers to determine which subset of the received channels should be selected. The selection layer will involve memory copy and increase the inference time. Therefore, to get an efficient network practically, a more general method is needed, which could take the network topology into consideration during pruning, so as to avoid additional post-processing.

Tackling this issue, we propose GAP for network topology-adaptive pruning. In the method, the network is viewed as a computational graph $G=\{V,E\}$, with the vertices $\{V\}$ denoting the computation operations, 
and the edges $\{E\}$ describing the information flow. 
In GAP, we conduct network pruning by removal of certain vertices or edges based on graph topology analysis. According to graph theory, the vertices can be divided into articulation points and non-articulation points, where an articulation point of graph $G$ is a vertex, of which removal disconnects $G$~\cite{cormen2009introduction}. To guarantee the information flow from the input to output, only the non-articulation points can be pruned. Similarly, the edges can be classified into bridges and non-bridges, while a bridge of $G$ is an edge, of which removal disconnects $G$. Only the non-bridges can be removed, otherwise the information flow will be broken off.

The whole procedure of GAP follows the framework of sparsity-induced weight pruning methods: 1) conduct structural sparsity constraint on the model parameters during training, 2) prune the vertices or edges with minor significance, 3) finetune the pruned graph. 
In GAP, pruning can be conducted at  either vertex-level or edge-level. At vertex-level, the graph topology is considered in order to avoid post-processing which may affect inference efficiency. 
For the vertices on the same cross-connection, regularization is conducted collaboratively using group sparsity to prune them all or keep them all.  
At edge-level, we mainly focus on the slimming of the multiple paths. Thus the graph is analyzed at coarser level.
Although a coaser-level pruning may suffer from more serious  performance degradation~\cite{mao2017exploring}, edge-level pruning is still considered in our method, since it not only reduces the computation cost but also the memory access times, which can further accelarate the inference in realtime.


For finetuning step, we introduce a self-taught KD procedure. In traditional KD methods, more complex networks are used as teachers to guide the student network training. In the proposed method, we choose to use the original model as the teacher. So it is actually a self-taught mechanism. 

The contribution of this work can be summarized as:



1.~We propose the GAP method for topology-adaptive CNN inference optimization, which does not need any post-processing even when the network contains cross-connections.

2. In GAP, a CNN model can be pruned at vertex-level as well as edge-level for the networks with multi-path data flow.

3. A self-taught KD mechanism for finetuning is proposed to further improve  performance of the pruned network.

\section{Related work}
The inference optimization methods can be categorized into two classes: 1) \emph{reducing the model representation precision} and 2) \emph{reducing the number of model parameters}. 

\textbf{1) Reducing the model representation precision.} This category includes network quantization and binarization. Network quantization compresses the bitwidth of the weights, activations or both~\cite{gong2014compressing,han2015deep}.  Extreme quantization is to binarize the network~\cite{hubara2016binarized}, using 1-bit to represent a value. Such kind of works using fix-point or binary representation need specially designed compute acceleration library or hardware. Additionally, the binarization methods always suffer from significant accuracy loss.


\textbf{2) Reducing the number of model parameters.} Such kind of methods 
can be categorized into new network architecture designs, TD, weight pruning and KD. 

\emph{New network architecture designs.} Some researches explore to get the inference-efficiency at the beginning of network design, such as 
SqueezeNet~\cite{iandola2016squeezenet} and MobileNet~\cite{howard2017mobilenets} 
\emph{etc}. The main technique is to replace the large convolution filters by a stack of small filters and train the network end-to-end. 

\emph{Tensor decomposition} aims to reduce the FLOPs by decomposing a large 4-D  filter into several small tensors by Canonical Polyadic (CP) decomposition~\cite{lebedev2014speeding} or Tucker decomposition~\cite{kim2015compression}. The TD-based methods will introduce more $1\times 1$ convolution layers, which is less cache-efficient. As a result, the practical speed-up ratio (SR) is not as ideal as the theoretical value. 

\emph{Weight pruning} can reduce the model size by removing some redundant parameters. 
The work in~\cite{han2015learning} employs the magnitude of the weights to evaluate the weight importance to determine which parameters should be removed. 
This kind of pruning (fine-grained pruning) needs dedicated compute libraries or/and hardware design, such as EIE~\cite{han2016eie}. 
More works explore to find structural pruning, which can get practical speed-up over existing compute libraries. 
The structural pruning methods \cite{he2017channel,wen2016learning} remove part of the filters/channels offline based on certain importance measurement or online using sparsity constraint while training.   
A finetuning procedure is conducted to compensate the performance loss. 
However, such 
methods usually ignore the network topology. As a result, additional post-processing layer may be needed to deal with complex network structure, such as cross-connections.

\emph{Knowledge Distillation} is proposed by Hinton \emph{et al.} to guide the student network training by a pretrained teacher model using soft target~\cite{hinton2015distilling}. The method aims to transfer the knowledge from a complex teacher model to the student network. FitNet~\cite{romero2014fitnets} extends the method by distilling the knowledge not only in the output but also the intermediate representations. 

In this paper, our method falls into the type of weight pruning based on structural sparsity constraint, and the strategy of KD is adopted to distill the knowledge from original model to maintain the model performance.

\section{Proposed method}
Given a pretrained model, graph pruning can be conducted using the following steps:

\noindent
1) Re-train with sparsity regularization. The sparsity is conducted on some parameters with certain structural pattern to make some vertices or edges removable;

\noindent
2) Sort all the weights and determine the pruning threshold;

\noindent
3) Remove the correpsonding vertices or edges according to the threshold;

\noindent
4) Finetune the pruned graph with or without self-taught KD.


\subsection{Notations}
In this section, we use two kinds of description to represent a CNN network: mathematics and graph.

\textbf{Mathematics.} For convolution, we use $\mathbf{X}$, $\mathbf{W}$, $\mathbf{Z}$ to denote the input feature maps, convolution kernels and output feature maps, respectively. 
Each channel $\mathbf{Z}_i$ in the output feature maps corresponds to a filter $\mathbf{W}_i$, and the batch normalized result is represented by $\mathbf{\hat{Z}}_i$,
\begin{equation}
\label{conv_channel}
\mathbf{Z}_i=\mathbf{X}*\mathbf{W}_i, \     \   \mathbf{\hat{Z}}_i=\frac{\mathbf{Z}_i-\mu_i}{\sqrt{\sigma_i^2+\epsilon}}\cdot \gamma_i + \beta_i,
\end{equation}
where $\mu_i$ and $\sigma_i^2$ are mean and variance of the channel, $\gamma_i$ and $\beta_i$ are the scaling factor and bias factor, respectively.

We use a symbol $W$ to represent all the parameters in CNN, including $\left\{\mathbf{W}\right\}$, $\left\{\boldsymbol{\gamma}\right\}$, $\left\{\boldsymbol{\beta}\right\}$ and also the other parameters, such as those in the FC layer. $W$ can be learned by the following optimization,
\begin{equation}
\label{originalCNN}
\min_{W} f\left(\mathcal{I}, W \right) + R(W),
\end{equation}
where, $\mathcal{I}$ denotes the input pairs including data and labels, $f\left(\cdot \right)$ is the loss function, $R\left(\cdot \right)$ is the regularization used in the training process, such as Frobenious-norm for weight decay.

\textbf{Graph.} We use a graph $G=\{V,E\}$ to represent a network, where vertices $\{V\}$ denote the computation operations and edges $\{E\}$ show the data flow. Subset module of DenseNet and ResNet~\cite{he2016deep} are shown in Figures \ref{DenseNet} and \ref{ResNet}.
In CNN, the computation operations include convolution, BN, activation, concat, add and FC, \emph{etc}. Since convolution accounts for the majority of computational load, we focus on the pruning of convolution vertices in our method. 

\subsection{Vertex-level pruning}
\label{subsec:vertex_slimming}
There are several rules to represent a graph for vertex-level pruning. 1) A convolution vertex represents a single filter rather than a set of filters in a convolution layer. Otherwise, we cannot prune the network at the fine filter-level when pruning  graph at vertex level. 
2) Similarly, a BN vertex represents the operation on a single channel. 
3) Because the activation functions are always placed after BN and are conducted element-wise, we fuse the activation function into the BN layer to simplify the graph illustration. 
In the following we will show 1) how the original channel-level pruning is performed~\cite{liu2017learning} and 2) how we perform the structural vertex-level pruning with consideration of the graph topology. 
\begin{figure}[!htb]
\vspace{-0.4cm}
\centering
	\begin{minipage}[b]{0.48\linewidth}
	\centerline{
		\includegraphics[width=\linewidth]{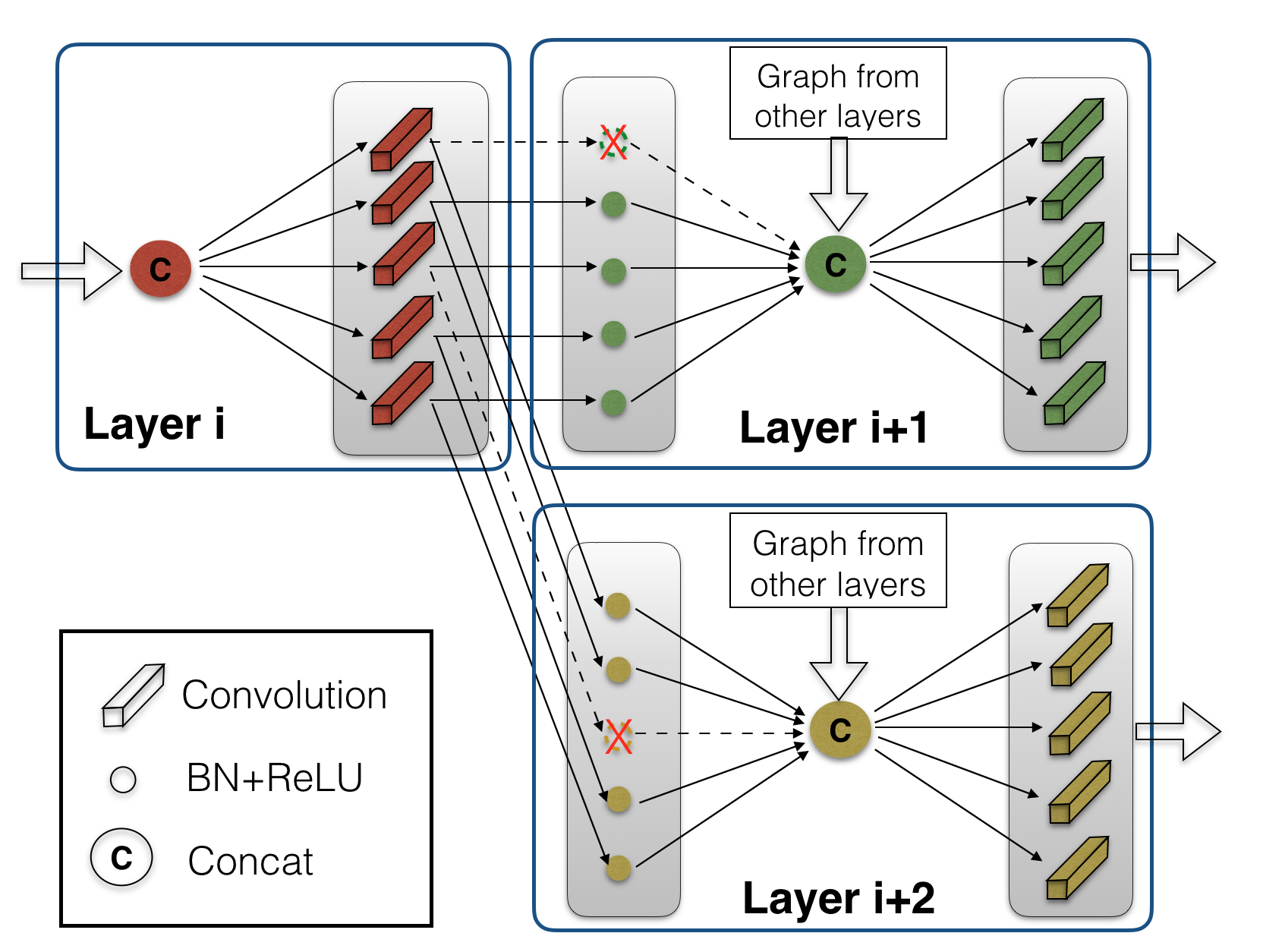}
	}
	\centerline{\footnotesize (a) channel-level pruning}
	\vspace{-0.2cm}
	\end{minipage}	
  	\hfill
	\begin{minipage}[b]{0.48\linewidth}
	\centerline{
		\includegraphics[width=\linewidth]{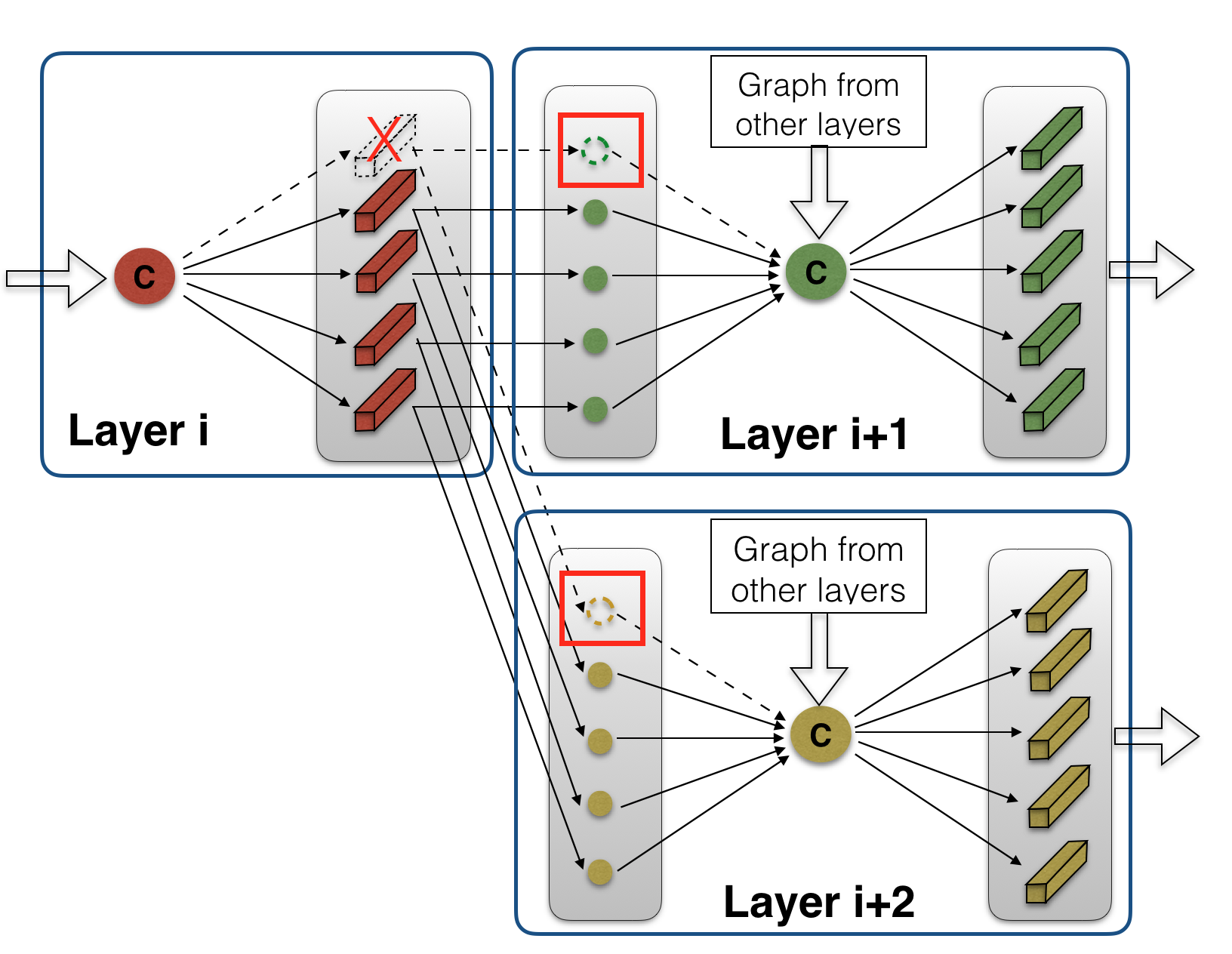}
	}
	\centerline{\footnotesize (b) vertex-level pruning}
	\vspace{-0.2cm}
	\end{minipage}	
	\caption{Pruning in DenseNet.}
	\label{DenseNet}
\end{figure}

\begin{figure}[!h]
\vspace{-0.7cm}
\centering
	\begin{minipage}[b]{0.48\linewidth}
	\centerline{
		\includegraphics[width=\linewidth]{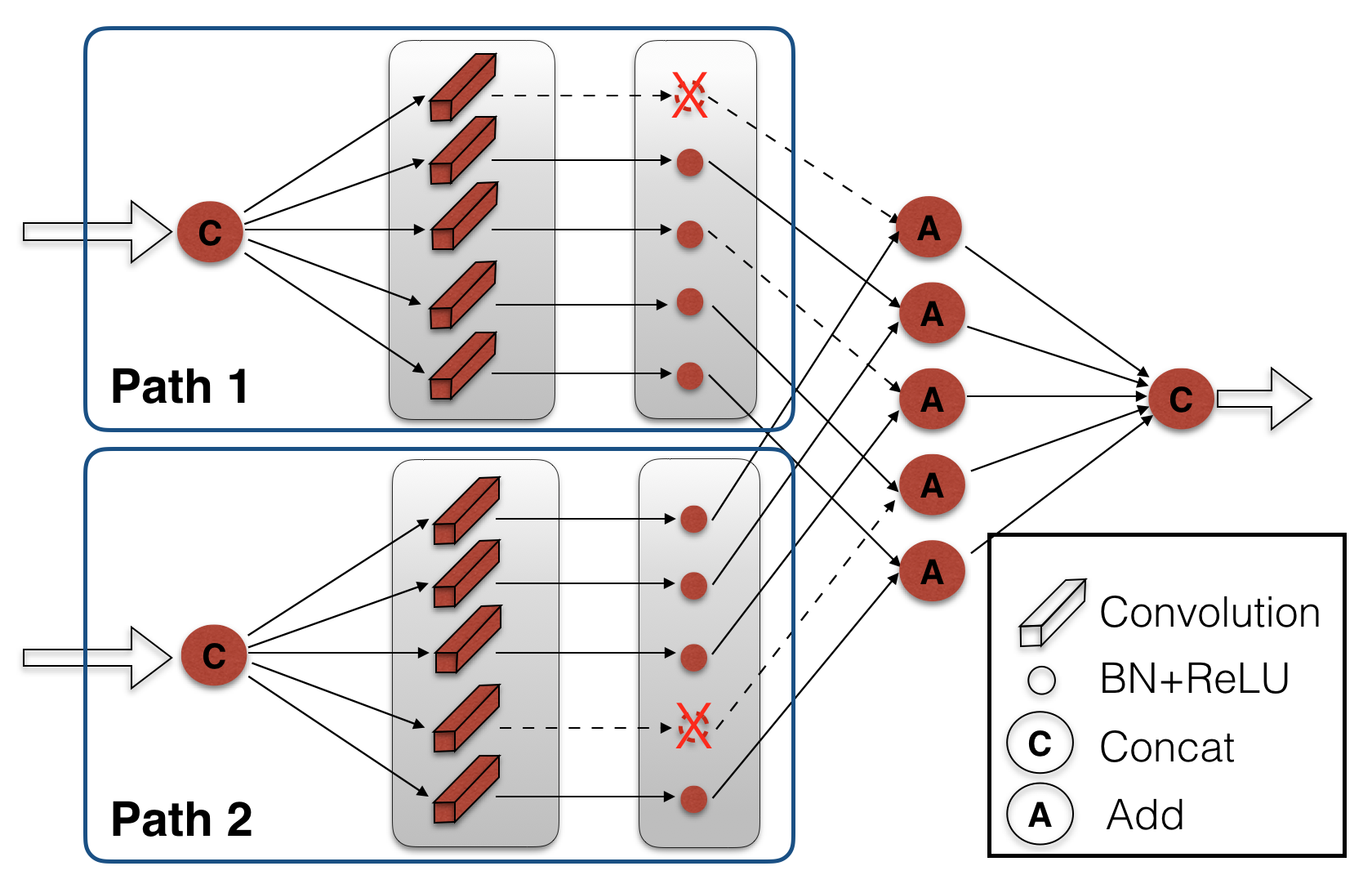}
	}
	\centerline{\footnotesize(a)}
	\vspace{-0.2cm}
	\end{minipage}	
  	\hfill
	\begin{minipage}[b]{0.48\linewidth}
	\centerline{
		\includegraphics[width=0.95\linewidth]{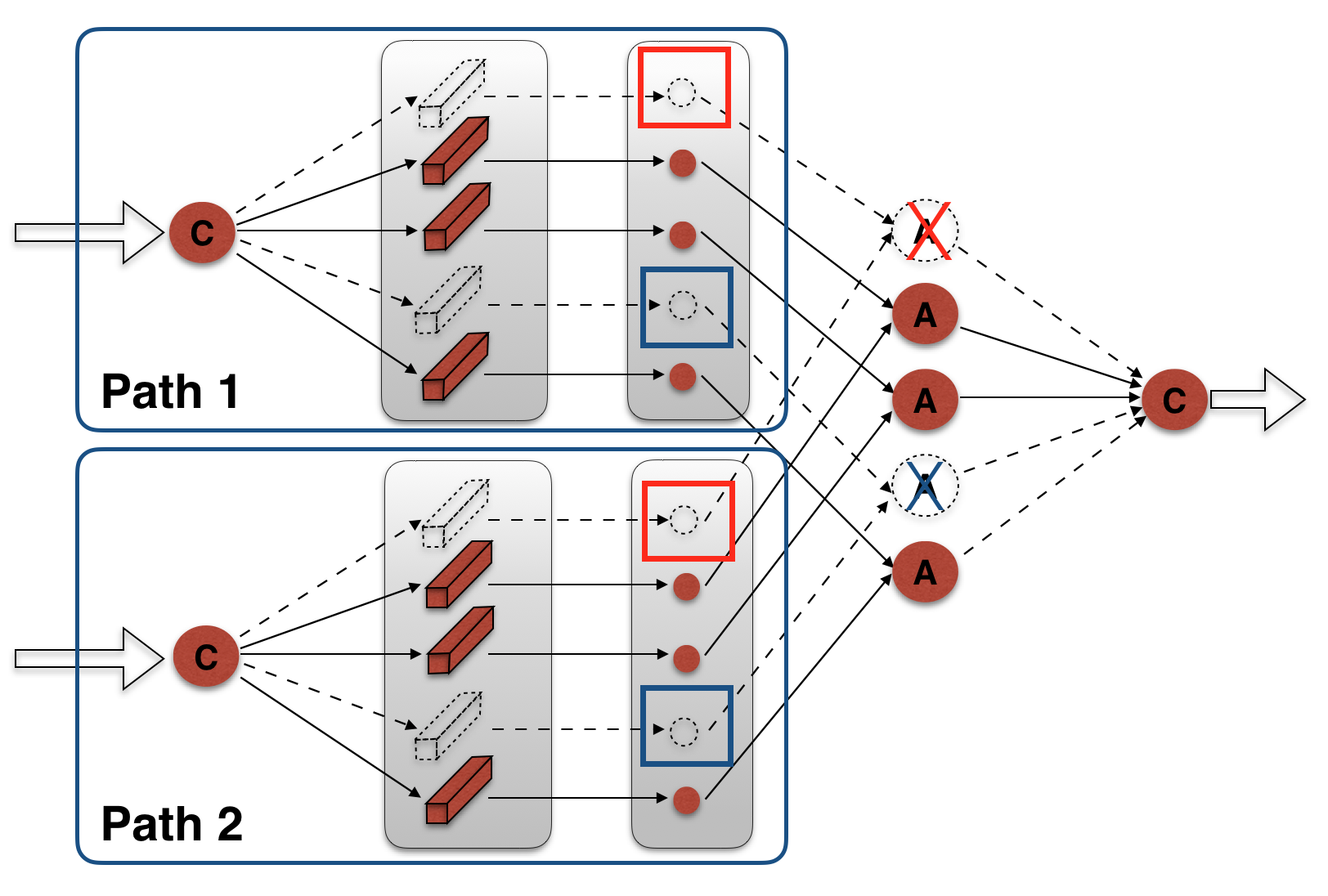}
	}
	\centerline{\footnotesize(b)}
	\vspace{-0.2cm}
	\end{minipage}		
	\caption{Pruning in residual module. (a) channel-level pruning; (b) vertex-level pruning.}
	\label{ResNet}
	\vspace{-0.2cm}
\end{figure}

For CNN with BN layers, 
the scaling factors in BN layers can play a role of measuring the importance of each channel, 
and thus can be directly used for channel selection with sparsity regularization. The channel-level pruning can be obtained by modifying the optimization in Eq.(\ref{originalCNN}) as
\begin{equation}
\label{channelNS}
\min_{W} f\left(\mathcal{I}, W \right) + R(W) + \lambda_s R_s\left(\left\{\boldsymbol{\gamma}\right\}\right),
\end{equation}
where $R_s\left(\cdot\right)$ is the sparsity regularization, which is typical realized using $\ell_1$-norm, $\lambda_s$ is the balance parameter which can trade-off between the sparsity loss and the original loss.

In a graph, channel-level pruning is targeted to remove the insignificant BN vertices. 
However, in DensNet as shown in Figure \ref{DenseNet}(a), removal of a BN vertex cannot result in removal of the preceding convolution vertex, because the convolution vertex still has outgoing edges connecting with other BN vertices. Similarly in residual module as shown in Figure \ref{ResNet}(a), 
there is n-to-1 connection due to the add-operation. The add-vertex cannot be removed if one of its incoming is remained, and once an add-vertex is remained, all its incoming edges should be remained to guarantee the validity of the data flow.

Therefore, to remove a certain convolution vertex, the graph topology should be taken into consideration, especially for the networks with cross-connections. Based on the conception, we propose to adaptively prune network at vertex-level by a more structural way. 

Firstly, the BN vertices are classified into articulation points $\{V_a\}$ and non-articulation points $\{\bar{V}_a\}$;

Secondly, $\{\bar{V}_a\}$ is further split into to 1-to-1 connection $\{\bar{V}_a^{(1\_1)}\}$, 1-to-n connection $\{\bar{V}_a^{(1\_n)}\}$ and n-to-1 connection $\{\bar{V}_a^{(n\_1)}\}$ BN vertices. If we use $\mathcal{P}_v$ and $\mathcal{C}_v$ to represent the set of the parent  and child vertices of $v$, then the definitions are as follows,

\begin{equation}
\vspace{-0.3cm}
\nonumber
 ~~~ \{\bar{V}_a^{(1\_n)}\}=\{v\in\bar{V}_a \mid \exists w\in \bar{V}_a, w\ne v, \mathcal{P}_v \cap \mathcal{P}_w \neq \emptyset\},
\end{equation}
\begin{equation}
\nonumber
 ~~~ \{\bar{V}_a^{(n\_1)}\}=\{ v\in\bar{V}_a \mid \exists w\in \bar{V}_a, w \ne v, \mathcal{C}_v\cap \mathcal{C}_w \neq \emptyset\},
\end{equation}
\begin{equation}
\begin{aligned}
 \{\bar{V}_a^{(1\_1)}\}=\{ v\in\bar{V}_a &\mid \forall w\in \bar{V}_a, w \ne v, \\ &\mathcal{C}_v \cap \mathcal{C}_w = \emptyset, \mathcal{P}_v \cap \mathcal{P}_w= \emptyset\},
\end{aligned}
\end{equation}

In GAP, 
we ignore cross-connections when the shared child vertex is ``concat''.  When different feature maps are combined through ``concat'', there is still no correlation among them. As a result, for the parent vertices of concat-vertex, whether they can be pruned still depends on themselves.

Finally, different constraints are conducted on different subsets,
\begin{equation}
\nonumber
\min_{W} f\left(\mathcal{I}, W \right) + R(W) + \lambda_s R_s\left(\left\{\boldsymbol{\gamma}_s\right\}\right) + \lambda_{gs} R_{gs}\left(\left\{\boldsymbol{\gamma}_{gs}\right\}\right)
\end{equation}
\begin{equation}
\label{kernelNS}
\text{s.t.}~~ \gamma_s \in \{\bar{V}_a^{(1\_1)}\}, ~~ \gamma_{gs} \in \{\bar{V}_a^{(1\_n)}\} \cup \{\bar{V}_a^{(n\_1)}\}
\end{equation}
The vertices in $\{\bar{V}_a^{(1\_1)}\}$ are regularized by $\ell_1$-norm $R_s\left(\cdot\right)$, and those in $\{\bar{V}_a^{(1\_n)}\}$ and $\{\bar{V}_a^{(n\_1)}\}$ are constrained by group sparsity, using $\ell_{2,1}$-norm $R_{gs}\left(\cdot\right)$, while each group denotes the vertices that share the same parent or child vertex.

\subsection{Edge-level pruning}
\begin{figure}[htb]
\centering
\vspace{-0.15cm}
	\begin{minipage}[b]{1.0\linewidth}
	\centerline{
		\includegraphics[width=0.7\linewidth]{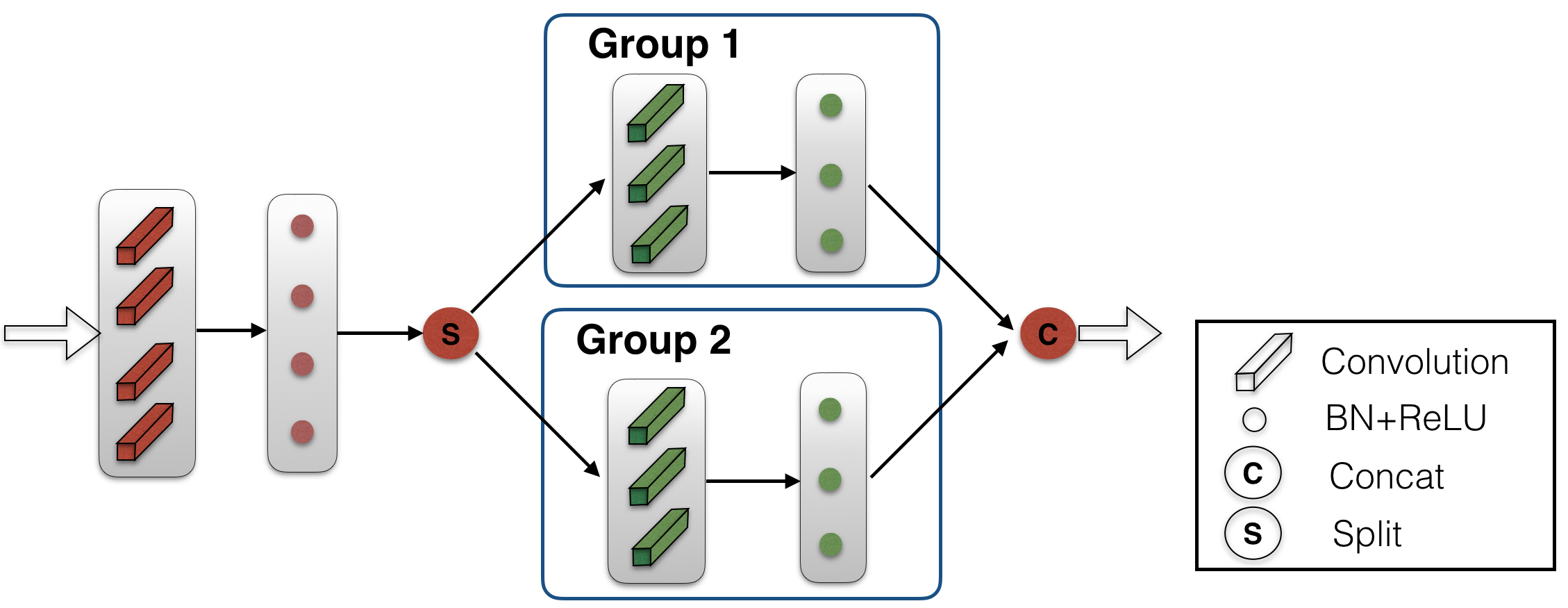}
	}
	\centerline{\footnotesize(a)}
	\end{minipage}
	\begin{minipage}[b]{0.45\linewidth}
	\centerline{
		\includegraphics[width=\linewidth]{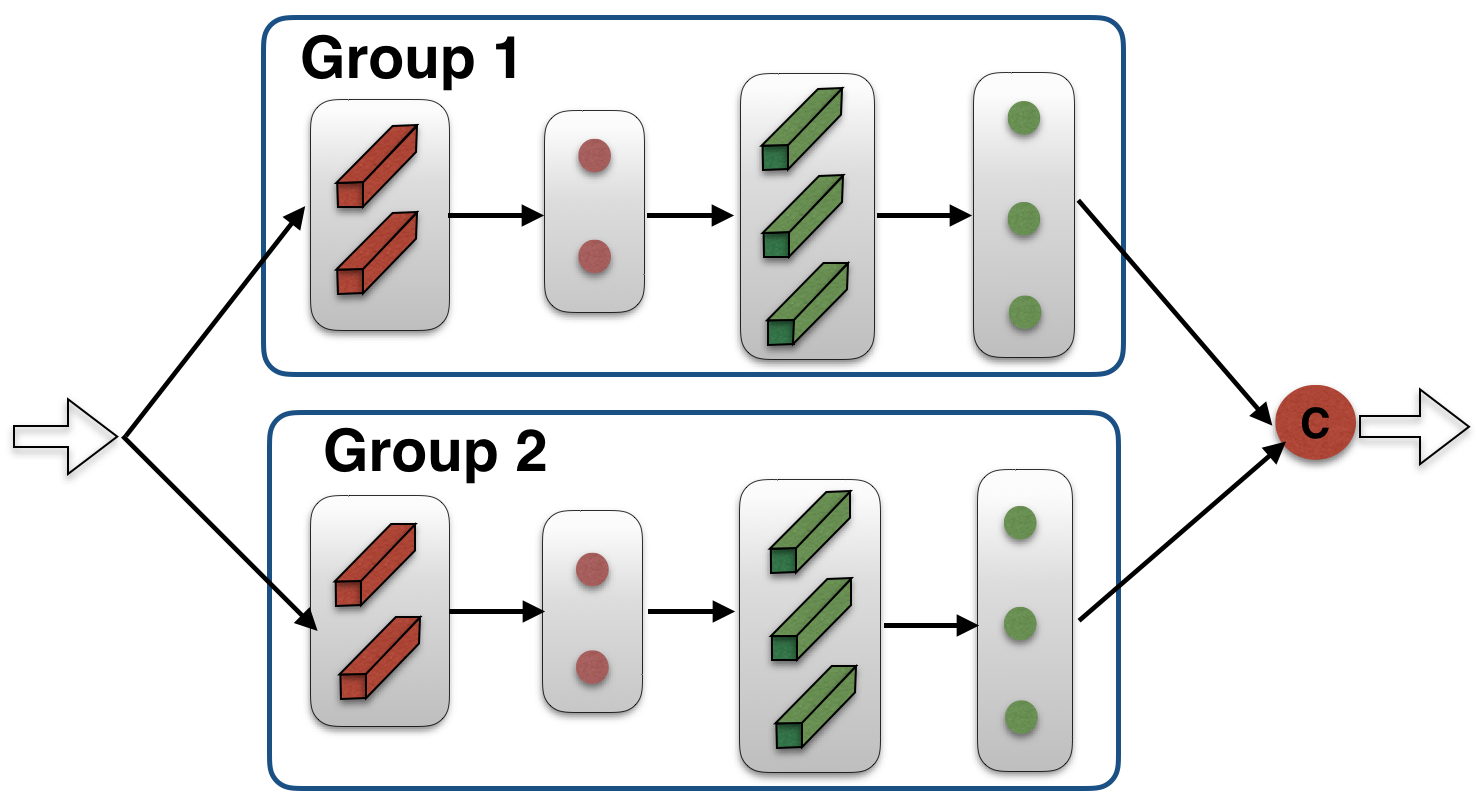}
	}
	\centerline{\footnotesize(b)}
	\vspace{-0.2cm}
	\end{minipage}	
  	\hspace{0.4cm}
	\begin{minipage}[b]{0.45\linewidth}
	\centerline{
		\includegraphics[width=\linewidth]{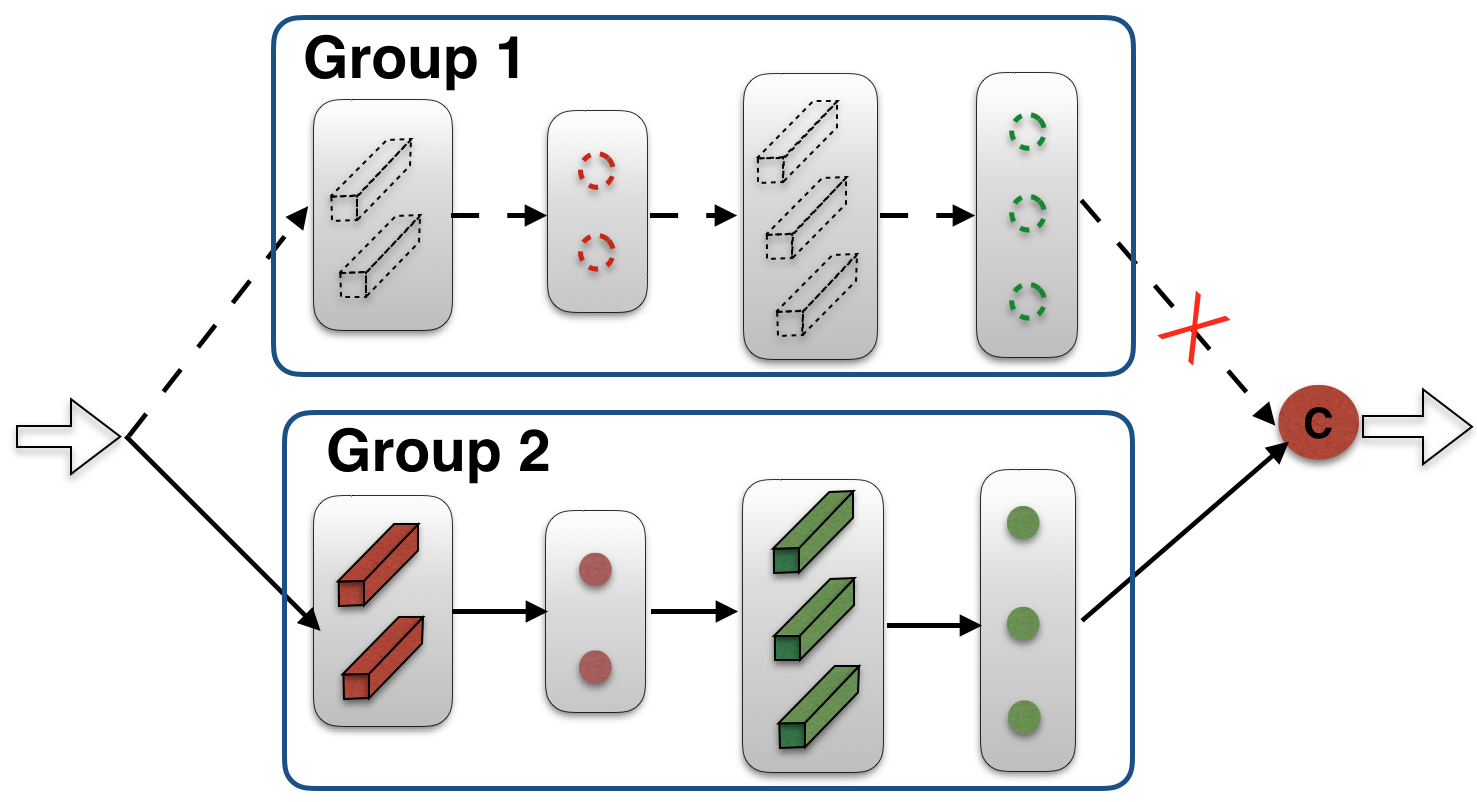}
	}
	\centerline{\footnotesize(c)}
	\vspace{-0.2cm}
	\end{minipage}		
	\caption{Edge-level pruning for group convolution. (a) group convolution original structure; (b) group convolution equivalent structure; (c) edge-level pruning of group convolution.}
	\label{ResNeXt_connection}
\end{figure}

\begin{figure}[htb]
\vspace{-0.4cm}
\centering
	\begin{minipage}[b]{0.43\linewidth}
	\centerline{
		\includegraphics[width=\linewidth]{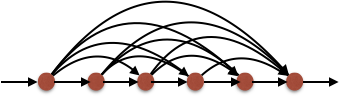}
	}
	\centerline{\footnotesize{(a)}}
	\vspace{-0.3cm}
	\end{minipage}	
  	\hspace{0.4cm}
	\begin{minipage}[b]{0.43\linewidth}
	\centerline{
		\includegraphics[width=\linewidth]{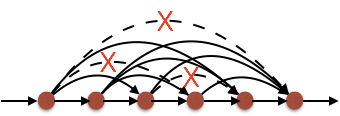}
	}
	\centerline{\footnotesize{(b)}}
	\vspace{-0.3cm}
	\end{minipage}	
	\caption{Edge-level pruning in DenseNet. (a) before pruning; (b) after pruning.}
	\label{Densenet_connection}
	\vspace{-0.3cm}
\end{figure}

In this section, we introduce the edge-level pruning. Recently, many networks are proposed using multi-path data flow, such as the inception module in GoogleNet, the fire module in SqueezeNet and the ``dense'' connection in DenseNet. Specifically, group convolution is a special case for multi-path design, with all paths identical. Figure \ref{ResNeXt_connection}(a) shows the original group convolution module, while (b) shows an equivalent structure, which is easier for topology analysis. As the network structure are usually designed with redundancy to help solving the highly non-convex optimization \cite{luo2017thinet}, not all paths are essential for the network performance. Thus certain paths can be pruned to reduce the model size and FLOPs of inference. 
In a graph, this can be realized by edge-level pruning. Different from Section \ref{subsec:vertex_slimming}, here we can treat a network as graph at a coarser level: a set of filters in a convolution layer is regarded as a single vertex. Similarly, a BN vertex represents a whole BN layer in edge-level pruning. When there are multiple paths for data flow, the edges on such paths become non-bridge. Thus the multi-path pruning is equivalent to removing part of the non-bridge edges. And the sparsity regularization to make non-bridge edges pruning is conducted as steps below:

Firstly, the non-bridge edges are selected as candidates to be pruned. As shown in Figures   \ref{ResNeXt_connection}(c) and \ref{Densenet_connection}(b), if one edge is removed, the whole path will be disabled. Thus there is no need to regulate all the non-bridge edges on a certain path. We only choose the last edge in each path to conduct pruning. Furthermore, in CNNs, multiple paths are always combined together using a ``concat'' operation. Therefore, we use concat-vertex to detect the edges to be pruned. The set of selected edges is denoted as $\{E_s\}$.
Secondly, each selected edge is scaled by an additional parameter $\gamma_e$, acting as a measurement of the edge's importance. The edge scaling factors are therefore constrained using sparsity regularization,
\begin{equation}
\label{connectionNS}
\min_{W} f\left(\mathcal{I}, W \right) + R(W) + \lambda_{es} R_{es}\left(\left\{\boldsymbol{\gamma_e}\right\}\right),
\end{equation}
where, $R_{es}\left(\cdot\right)$ denotes $\ell_1$-norm on the scaling factors $\left\{\boldsymbol{\gamma_e}\right\}$, $\lambda_{es}$ is the balance parameter.

\subsection{Self-taught KD}
After training with sparsity constraints, all the scaling factors can be sorted. The vertices or edges with scaling factor smaller than a certain threshold are pruned, while the threshold can be obtained by the pruning ratio. The pruned graph may suffer from certain  performance degradation, and it can be compensated by finetuning. 
In addition to naive finetuning, we propose to finetune the network using a self-taught KD strategy in this paper. 


For the pruned network, the original model is apparently a more complex model with better performance, and it can act as the teacher in KD for finetuning. In addition, the pretrained model is already provided, 
which is rather important in practice, 
as there is always limited resource and time to train a more complex teacher model for a specific task.
As the knowledge is distilled from the original model to the pruned network, we denote it as self-taught KD. In the following sections, experimental results will show that the performance of the pruned network is sufficiently improved when compared with the naive finetuning strategy.

\section{Experiments}
\subsection{Implementations}
\begin{figure*}[!htb]
\vspace{-0.6cm}
\centering
	\begin{minipage}[b]{0.27\linewidth}
	\centerline{
		\includegraphics[width=\linewidth]{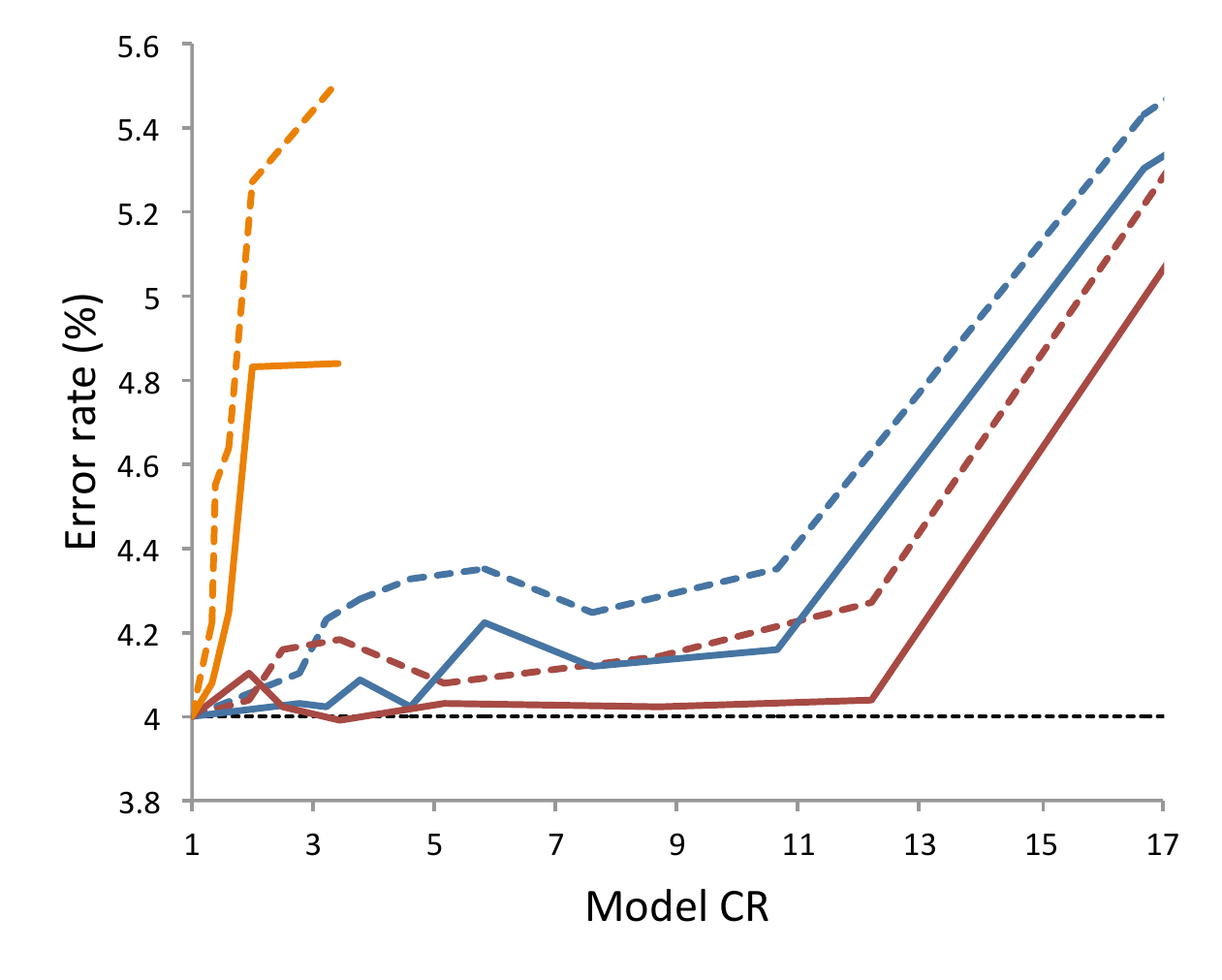}
	}
	\vspace{-0.3cm}
	\end{minipage}	
  	\hfill
	\begin{minipage}[b]{0.27\linewidth}
	\centerline{
		\includegraphics[width=\linewidth]{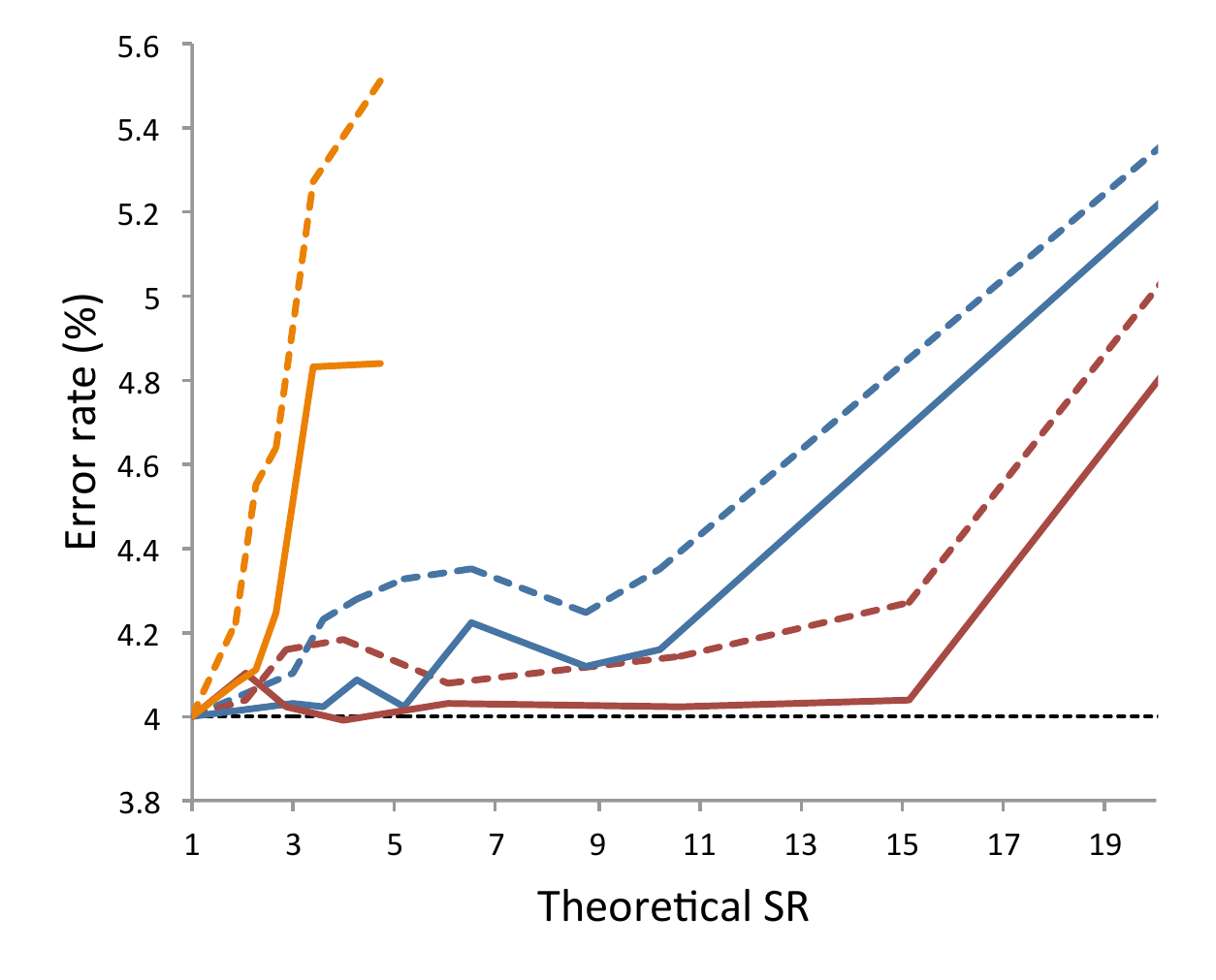}
	}
	\vspace{-0.3cm}
	\end{minipage}	
	\hfill
	\begin{minipage}[b]{0.38\linewidth}
	\centerline{
		\includegraphics[width=\linewidth]{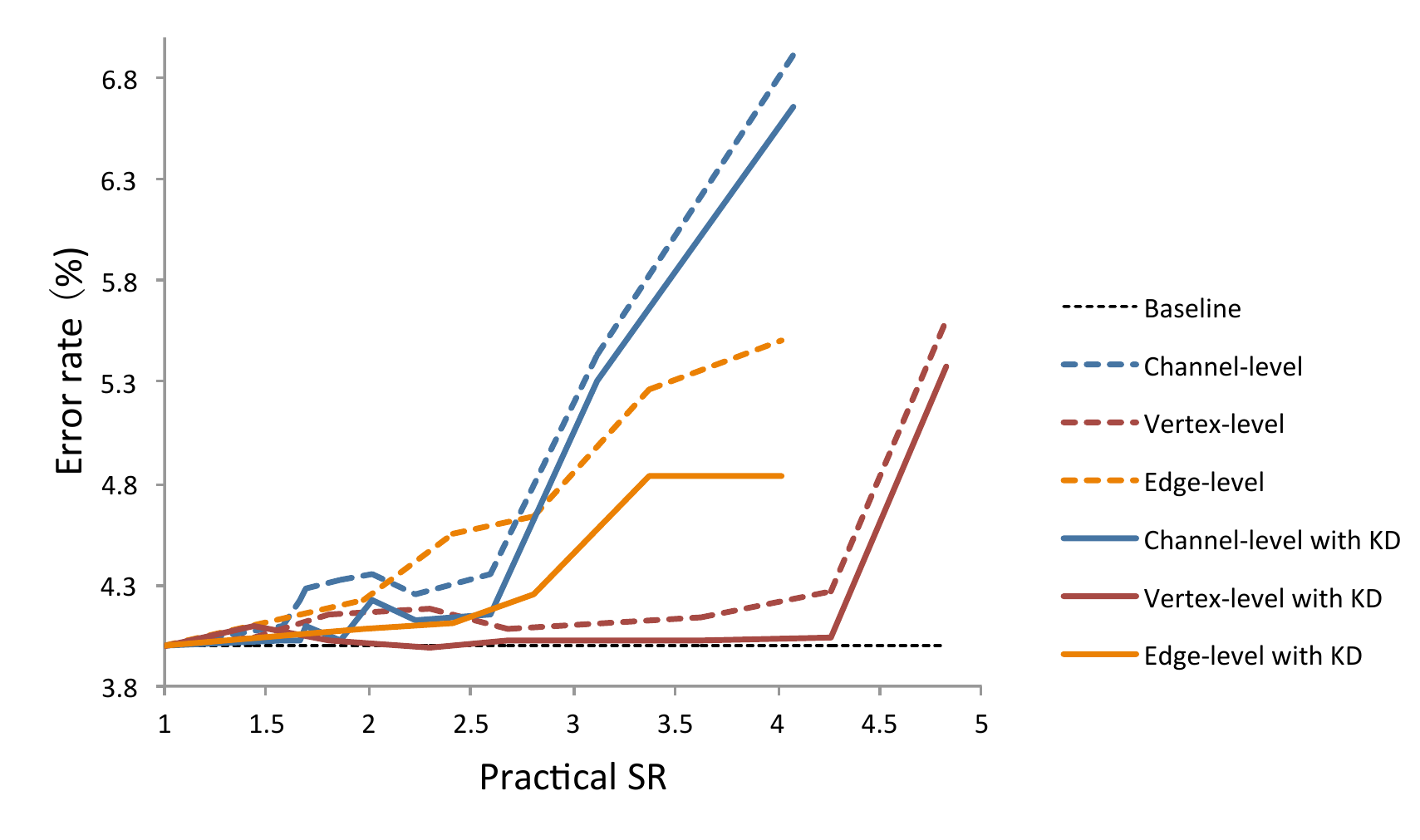}
	}
	\vspace{-0.3cm}
	\end{minipage}
	\caption{Network performance with respect to model CR, theoretical SR and practical SR for pruning ResNeXt-29 on CIFAR10. Dashed lines show the results by  finetuning without KD and the solid ones denote results through finetuning with self-taught KD. }
	\label{ResNeXt_quantitative}
\end{figure*}

\begin{table*}[!htbp]
\vspace{-0.2cm}
\caption{Pruning results on CIFAR10.}
\label{ResNeXt_CIFAR10}
\vspace{-0.2cm}
\small
\renewcommand{\arraystretch}{0.9}
\centering
\begin{tabular}{ c| c |c |c |c |c |c |c |c |c }
\hline
& & \tabincell{c}{Pruned \\ /\%} & \tabincell{c}{Error /\%\\\scriptsize{(w/o KD)}} & \tabincell{c}{Error /\%\\ \scriptsize{(w/ KD)}} & \tabincell{c}{Model Size\\\scriptsize{(MB)}} &\tabincell{c}{Model\\ CR}& FLOPs & \tabincell{c}{Theoretical\\SR}& \tabincell{c}{Practical\\SR}\\
\hline \hline
\multirow{4}{*}{ResNeXt-29} & Baseline (Our impl.) & - & 4.00 & - & 34.43 & - & 5.00G & - & - \\
\cline{2-10}
&Channel-level\tiny{~\cite{liu2017learning}} & 60\% & 4.28 & 4.09 & 9.10 & 3.78 & 1.18G & 4.24 & 1.69\\
&Vertex-level & 60\% & 4.08 & 4.03 & 6.71 & 5.13 & 0.83G & 6.02 & 2.68 \\
&Edge-level & 60\% & 4.55 & 4.11 & 24.61 & 1.40 & 2.23G& 2.24 & 2.41 \\
\hline \hline
\multirow{4}{*}{DenseNet-40} & Baseline (Our impl.) & - & 5.60 & - & 1.06 & - & 288M & - & - \\
\cline{2-10}
& Channel-level\tiny{~\cite{liu2017learning}} & 50\% & 6.38 & 5.70 & 0.58 & 1.83 & 185M & 1.56 & 1.03 \\
& Vertex-level & 50\% & 6.14 & 5.67 & 0.52 & 2.04 & 138M & 2.09 & 1.27 \\
& Edge-level & 50\% & 6.24 & 6.00 & 0.90 & 1.18 & 207M & 1.39 & 1.11 \\
\hline \hline
\multirow{3}{*}{ResNet-164} & Baseline (Our impl.) & - & 5.11 & - & 1.70 & - & 251.0M & - & - \\
\cline{2-10}
&Channel-level\tiny{~\cite{liu2017learning}} & 75\% & 5.50 & 5.32 & 0.71 & 2.39 & 71.3M & 3.52 & 1.24 \\
&Vertex-level & 50\% & 5.47 & 5.36 & 0.71 & 2.39 & 70.1M & 3.58 & 1.48 \\
\hline
\end{tabular}
\vspace{-0.3cm}
\end{table*}

We evaluated the effectiveness of the proposed pruning method using two widely used datasets: CIFAR10~\cite{krizhevsky2009learning} and ImageNet LSVRC 2012~\cite{russakovsky2015imagenet}. Considering the ``topology-adaptive'' attribute of GAP, ResNet, DenseNet and ResNeXt~\cite{xie2017aggregated} were chosen for the evaluation. 

To evaluate the inference efficiency, we used three criteria: model compression ratio (CR), theoretical SR and practical SR. Model size and FLOPs before and after pruning were used to compute the model CR and theoretical SR. We used the practical SR as an additional indicator of inference efficiency, since the memory access and movement time are not considered in FLOPs. TensorFlow is used as the basic framework. The practical SR was evaluated with the library of CUDA8, cuDNN5 on a GPU (GTX1080Ti). As for all the network trainings, SGD with a Nesterov momentum of 0.9 was used as the optimizer. Weight decay for the networks was set to $10^{-4}$. For the sparsity regularization, the balance parameters for vertex-level were set to $\lambda_{gs}=10*\lambda_{s}$ for simplicity, as the structural group sparsity is harder to be sparsified. $\lambda_s$ was selected in  $\{10^{-2},10^{-3},10^{-4}\}$, based on sparseness of the targeted weights. For edge-level, $\lambda_{es}$ was searched in $\{10^{-1},10^{-2},10^{-3}\}$. All the layers were pruned simultaneously based on an adaptive threshold, which was determined by the pruning proportion.

\subsection{Experiments on CIFAR10}
\label{subsec:CIFAR10}

Data augmentation of CIFAR10 for training were conducted using random cropping and mirroring. Images were normalized channel-wise based on statistical values. For experiments on CIFAR10, ResNet-164, DenseNet-40 ($k$$=$$12$) and ResNeXt-29 ($8\times 64$d) were adopted. The original pre-trained models were implemented by ourselves based on TensorFlow, with the same settings as the authors'. In the first step, the pre-trained models were retrained with sparsity constraint using mini-batch size 128 for 10 epochs, with a learning rate of 0.01. All layers in the graph were pruned together and then finetuned for 90 epochs. Initial learning rate for finetuning was set to 0.01 and divided by 10 at 2/3 of the total epochs.
For self-taught KD, a temperature of $T=5$ was used and the relative weight for soft target was set to 1.

\begin{table*}[!htbp]
\vspace{-0.2cm}
\caption{Pruning results on ImageNet}
\label{ResNeXt_imagenet}
\vspace{-0.2cm}
\small
\renewcommand{\arraystretch}{0.9}
\centering
\begin{tabular}{ c| c |c |c |c |c |c |c |c |c }
\hline
& & \tabincell{c}{Pruned \\ /\%} & \tabincell{c}{Error /\%\\\scriptsize{(w/o KD)}} & \tabincell{c}{Error /\%\\ \scriptsize{(w/ KD)}} & \tabincell{c}{Model Size\\\scriptsize{(MB)}} &\tabincell{c}{Model\\ CR}& \tabincell{c}{FLOPs\\ \scriptsize{($\times 10^9$)}}& \tabincell{c}{Theoretical\\SR}& \tabincell{c}{Practical\\SR}\\
\hline \hline
\multirow{4}{*}{ResNeXt-50} & Baseline (Our impl.) & - & 24.47 & - & 25.03 & - & 4.26 & - & - \\
\cline{2-10}
& Channel-level\tiny{~\cite{liu2017learning}} & 50\% & 25.22 & 24.67 & 17.99 & 1.39 & 2.39 & 1.78 & 1.27 \\
& Vertex-level & 40\% & 26.15 & 24.44 & 14.99 & 1.67 & 2.29 & 1.86 & 1.51 \\
& Edge-level & 50\% & 26.19 & 25.27 & 16.88 & 1.48 & 2.37 & 1.80 & 1.48 \\
\hline \hline
\multirow{4}{*}{DenseNet-121} & Baseline (Our impl.) & - & 25.17 & - & 7.98 & - & 2.87 & - & - \\
\cline{2-10}
& Channel-level\tiny{~\cite{liu2017learning}} & 50\% & 25.50 & 25.19 & 4.74 & 1.68 & 2.06 & 1.39 & 1.01 \\
& Vertex-level & 20\% & 26.24 & 25.30 & 6.06 & 1.32 & 1.99 & 1.44 & 1.14 \\
& Edge-level & 50\% & 25.63 & 25.27 & 6.72 & 1.19 & 2.48 & 1.16 & 1.25 \\
\hline
\end{tabular}
\vspace{-0.1cm}
\end{table*}

\begin{figure*}[!htb]
\centering
	\begin{minipage}[b]{0.23\linewidth}
	\centerline{
		\includegraphics[width=\linewidth]{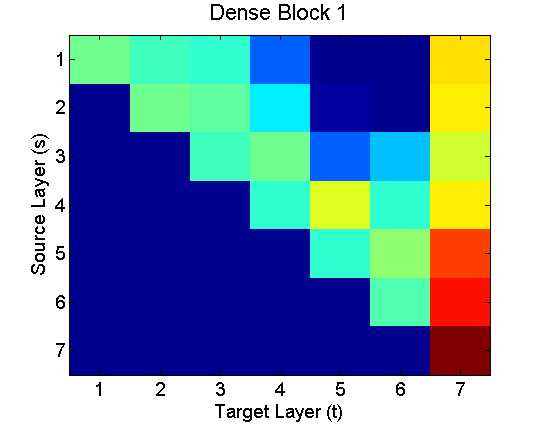}
	}
	\end{minipage}	
  	\hfill
	\begin{minipage}[b]{0.23\linewidth}
	\centerline{
		\includegraphics[width=\linewidth]{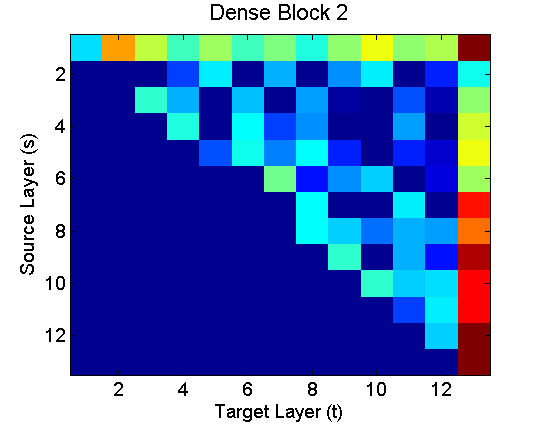}
	}
	\end{minipage}	
	\hfill
	\begin{minipage}[b]{0.23\linewidth}
	\centerline{
		\includegraphics[width=\linewidth]{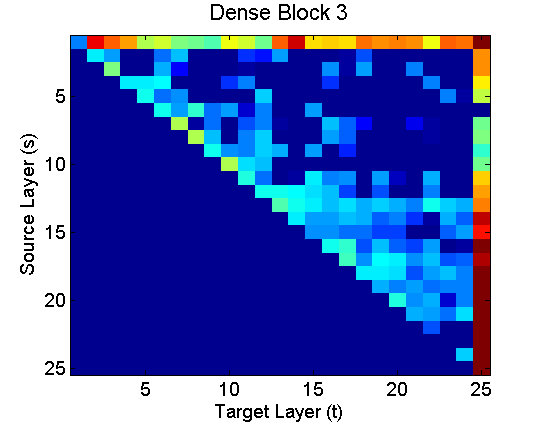}
	}
	\end{minipage}	
  	\hfill
	\begin{minipage}[b]{0.23\linewidth}
	\centerline{
		\includegraphics[width=\linewidth]{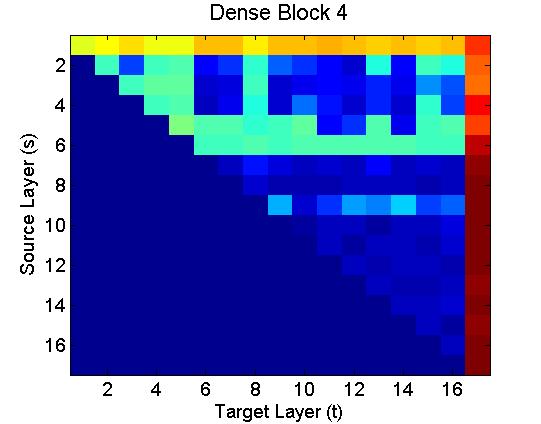}
	}
	\end{minipage}	
	\begin{minipage}[b]{0.033\linewidth}
	\centerline{
		\includegraphics[width=\linewidth]{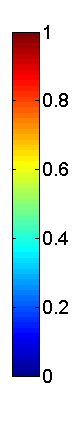}
	}
	\end{minipage}	
  	\hfill
\vspace{-0.2cm}
	\caption{The distribution of scaling parameters $\{\gamma_{e}\}$ after training with sparsity for edge-level pruning in DenseNet-121. The color of pixel $(s, t)~(s<t)$ encodes the $\{\gamma_e\}$ of the edge from layer $s$ to layer $t$. There are 4 blocks in DenseNet-121, each has 6, 12, 24 and 16 layers, respectively.  The last column of each block figure represents edges connected to the transition layer after each block.
	 }
	\label{densenet_edge}
	\vspace{-0.4cm}
\end{figure*}

Pruning results are shown in Table \ref{ResNeXt_CIFAR10}. In DenseNet and ResNeXt, the networks are pruned with the same percentage for channel-level, vertex-level and edge-level.
The results suggest that the strategy of finetuning with self-taught KD performs better than naive finetuning in restoring the degradation of classification accuracy caused by pruning.
By comparison, we can see that structurally pruning at vertex-level can get higher model CR and theoretical SR, while vertex-level pruning can still get better performance in classification error rate. In ResNeXt, with approximately no loss of accuracy, pruning 60\% off at vertex-level can get 2.68$\times$ practical SR while only {1.69$\times$} at channel-level. In DenseNet, channel-level pruning achieves almost no speed-up as it introduces additional selection operations, which increases the memory access time. Compared with ResNeXt, DenseNet is already a quite compact network, which is more difficult to be pruned. However, we still achieve 2.04$\times$ model CR and {1.27$\times$} practical SR through vertex-level pruning with marginal loss of performance. In ResNet, only channel-level and vertex-level pruning were conducted. The model was pruned to comparable model size and FLOPs at the two levels. At vertex-level, we can get 1.48$\times$ practical SR and 2.39$\times$ model CR with minor loss of accuracy.

Edge-level pruning leads to the largest remaining model size and FLOPs, because edge-level pruning can only prune part of the graph. 
In ResNeXt, only the edges contained in the group convolution can be pruned. Similarly,  only the dense connections can be removed in DenseNet. Additionally, edge-level gets the worst error rate. This is naturally because it prunes the network at a coarse-grained level, which will do more harm on the network \cite{mao2017exploring}. However, the benefit of edge-level pruning is that it has little gap between practical SR and theoretical SR: edge-level pruning reduces the number of computation operations, as a result it can reduce the FLOPs as well as the memory access times, while 
the theoretical SR ignores the issue of memory access. Specifically, in ResNeXt, the practical SR actually exceeds the theoretical one.

Furthermore, we use ResNeXt-29 to quantitatively analyze the network performance with respect to different model CRs, theoretical and practical SRs. Results are shown in Figure \ref{ResNeXt_quantitative}. Vertex-level pruning achieves lower error rate than channel-level with the same model CR, theoretical or practical SR, especially in the practical SR measurement. 
Through vertex-level pruning with self-taught KD finetuning, we can get approximately {12$\times$} model CR, {15$\times$} theoretical SR and {4.3$\times$} practical SR with nearly no loss of accuracy.  Although, edge-level suffers more in accuracy loss, it achieves larger practical SR at the same level of theoretical SR.  As shown in Figure \ref{ResNeXt_quantitative}(c), at the same error rate, edge-level pruning gets higher actual  speed-up than channel-level pruning. Finally, we can see that the performance is sufficiently improved by finetuning with the self-taught KD compared with naive finetuning.

\subsection{Experiments on ImageNet}
We adopted the same data augmentation scheme as in \cite{huang2016densely} for ImageNet. Top-1 error rate of a single center crop was used as the performance measurement. In the first step, the pretrained models were retrained with sparsity constraint using a mini-batch size of 256 on 4 GPUs for 1 epoch, with the learning rate being 0.01. For finetuning, the model was trained for 40 epochs, with an initial learning rate of 0.01, which was decreased by a factor of 10 at 15th and 30th epochs, respectively. For KD, $T=5$ was used and the relative weight for soft target was set to 100  making the loss magnitude of soft target and hard target comparable.

On ImageNet, ResNeXt-50 ($32\times 4$d) and DenseNet-BC-121 were validated, while the model settings were the same as in \cite{xie2017aggregated} and \cite{huang2016densely}. Table \ref{ResNeXt_imagenet} shows the pruning results. To better evaluate the performance, the models at channel-level and vertex-level are pruned to comparable model size and FLOPs. 
At vertex-level, the pruned networks can achieve quite similar error rates after finetuning with self-taught KD. At the same time, we get a model CR of 1.67$\times$ and practical SR of {1.51$\times$} in ResNeXt while 1.32$\times$ CR  and 1.14$\times$ practical SR in DenseNet.

\begin{figure}[!htb]
\vspace{-0.4cm}
\centering
	\centerline{
		\includegraphics[width=0.9\linewidth]{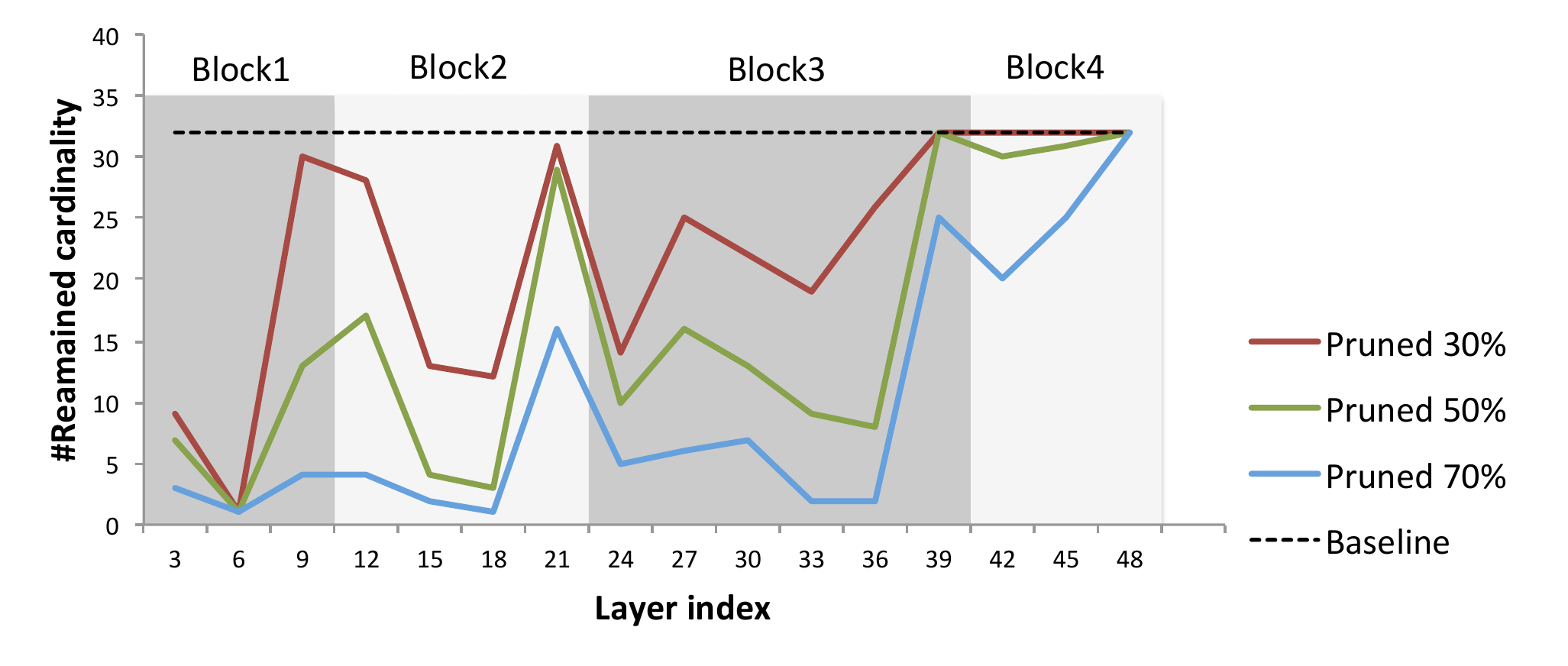}
	}
	\vspace{-0.4cm}
	\caption{Number of remained edges in ResNeXt-50 (32 $\times$ 4d) with different pruning percentages. The horizontal axis represents the index of group convolutional layers, vertical axis is the number of the remained cardinality (baseline is 32) in each group convolution. }
	\label{ResNeXt_remained_edges}
	\vspace{-0.1cm}
\end{figure}

For edge-level pruning, Figure~\ref{densenet_edge} shows the distribution of edge scaling parameters $\{\gamma_{e}\}$ after training with sparsity in DenseNet-121. 
$\{\gamma_{e}\}$ are all initialized to be 1.0,  and we can see that after training, non-zero $\gamma_{e}$ becomes sparse so that we can prune the edges with low scaling values. Furthermore, it can be observed that the information flow between blocks are critical, as indicated in the first row and last column of each block, while the layer connections within a block may have high redundancy.
For ResNeXt-50, Figure~\ref{ResNeXt_remained_edges} shows the number of remained edges in ResNeXt-50 with different pruning percentages. We use a global threshold to adaptively prune each layer, as the redundency may vary with different layers.


Because of the bottleneck structure in DenseNet-BC, edge-level pruning can only remove the convolutional layers with kernel size $1\times 1$. Therefore, it can only achieve a quite low model CR and FLOPs reduction. As for ResNeXt, only edges involved in the group convolutions can be pruned, thus the model CR and theoretical SR are also not high enough. However, as described in Section \ref{subsec:CIFAR10}, the benefit of edge-level pruning is that it can also reduce the memory access times and thus further accelerate the inference. In ResNeXt, the practical SR exceeds the theoretical value, and in DenseNet it has quite little gap with the theoretical SR.

\vspace{-0.5cm}
\section{Conclusion}
In this paper, we propose the GAP for CNN model compression and inference acceleration. By adaptive analysis of the graph, the method can directly remove certain vertices or edges to achieve a compact and efficient graph for inference by maintaining the original graph topology. 
The pruned network can achieve practical speed-up without any post-processing to deal with complex structures.
For finetuning, 
we adopt a self-taught KD strategy to improve the network performance. 
The strategy can sufficiently improve the model performance and it does not introduce extra workload,  
which is quite applicable for practical tasks.  Experimental results show it can make the inference more efficient, with high model CR and practical SR, while keeping the network performance very close to original model.
As the future work, we will develop an auto-tuning mechanism to search optimal hyper-parameters involved in the framework, and we are going to investigate the scheme to combine the vertex-level and edge-level pruning, so that a more rational mixed-level pruning can be conducted for a  network given computation resource or latency limitation.

\bibliographystyle{named}
\clearpage
\newpage
\bibliography{ref}

\end{document}